\documentclass[runningheads]{llncs}

\usepackage{hyperref}       
\usepackage{url}            
\usepackage{booktabs}       
\usepackage{amsfonts}       
\usepackage{nicefrac}       
\usepackage{microtype}      
\usepackage{xcolor}         
\usepackage{graphicx}
\usepackage{amsmath}
\begin{document}
\title{Active Inference in Hebbian Learning Networks} 
\titlerunning{Active Inference in Hebbian Learning Networks}


\author{Ali Safa\inst{1,2,3} \and
Tim Verbelen\inst{4} \and
Lars Keuninckx\inst{2} \and
 Ilja Ocket\inst{2} \and
André Bourdoux\inst{2} \and
 Francky Catthoor\inst{1,2} \and
Georges Gielen\inst{1,2} \and
 Gert Cauwenberghs\inst{3}}
 \authorrunning{A. Safa et al.}
 \institute{imec, Leuven, Belgium \and ESAT, KU Leuven, Belgium \and University of California at San Diego, La Jolla, USA \and VERSES Research Lab, Los Angeles, California, USA\\
 \email{Ali.Safa@imec.be}}

\maketitle              
\begin{abstract}
This work studies how brain-inspired neural ensembles equipped with local Hebbian plasticity can perform active inference (AIF) in order to control dynamical agents. A generative model capturing the environment dynamics is learned by a network composed of two distinct Hebbian ensembles: a \textit{posterior} network, which infers latent states given the observations, and a \textit{state transition} network, which predicts the next expected latent state given current state-action pairs. 
Experimental studies are conducted using the Mountain Car environment from the OpenAI gym suite, to study the effect of the various Hebbian network parameters on the task performance. It is shown that the proposed Hebbian AIF approach outperforms the use of Q-learning, while \textit{not requiring} any replay buffer, as in typical reinforcement learning systems. These results motivate further investigations of Hebbian learning for the design of AIF networks that can learn environment dynamics without the need for revisiting past buffered experiences.


\keywords{Active Inference  \and Hebbian Learning \and Sparse Coding.}
\end{abstract}

\section{Introduction}
\label{intro}
The study of Sparse Coding \cite{olhausend}, \cite{langev1}, \cite{NIPS20062d71b2ae}, \cite{sdparsecode} and Predictive Coding \cite{fristonpredic}, \cite{predcode2}, \cite{predcode} networks has gained much attention for understanding the mechanisms underlying learning and inference in the brain \cite{Olshausen1996}. In particular, it has been shown that the learning of the \textit{weight dictionary} used to project the input signals into sparse codes can be conducted via the biologically-plausible Hebbian learning mechanism \cite{9892362}, with experimental evidence behind this mechanism observed in the brain \cite{biandpoo}, \cite{Rao99}. Hebbian learning differs from the widely-used \textit{back-propagation of error} (backprop) technique due to its \textit{local} nature \cite{predcode}, \cite{stdp2}, \cite{contrsdiv}, where the weight $w_j$ of neuron $i$ is modified via a combination $f$ of the weight's input $x_j$ and the neuron's output $y_i$ (with $\eta_d$ the learning rate parameter):
\begin{equation}
     w_j \xleftarrow[]{} w_j + \eta_d f(y_i,x_j)
     \label{hebbianrule}
\end{equation}
When applied to layers that embark some form of competition between their neurons, the Hebbian mechanism in (\ref{hebbianrule}) leads to the \textit{unsupervised} learning of complementary features from the input signals \cite{doi:10.1073/pnas.1820458116}. 

At the same time, Active Inference (AIF) has gained huge interest as a \textit{first-principle} theory, explaining how biological agents evolve and perform actions in their environment \cite{activinfbook}, \cite{spikeaif}. In recent years, the use of deep neural networks (DNNs) for parameterizing generative models has gained much attention in AIF research \cite{ozan}, \cite{dai}, \cite{deepactinf2}. Deep AIF systems are typically composed of a \textit{posterior} network $q_{\Phi_P}(s_l|o_{l-1},a_{l-1})$, inferring the latent state $s_l$ given an incoming observation-action pair $\{o_{l-1},a_{l-1}\}$, and a \textit{state-transition} network $p_{\Phi_S}(s_l|s_{l-1}, a_{l-1})$, predicting the next latent state $s_l$ given the current state-action pair $\{s_{l-1},a_{l-1}\}$ \cite{ozan}. The state-transition network is used to generate the agent's roll-outs for different policies in order to compute the Expected Free Energy associated to each policy \cite{ozan}. Finally, a \textit{likelihood} network $p_{\Phi_L}(o_l|s_l)$ reconstructing the input observation $o_l$ from the latent state $s_l$ can also be implemented \cite{toonpaper}. Each network parameterizes its respective density function through weight tensors $\Phi_P, \Phi_S$ and $\Phi_L$.

In this work, we aim to study how AIF can be performed in Hebbian learning networks \textit{without resorting to backprop} (as typically used in deep AIF systems). 
Experiments conducted in the OpenAI Mountain Car environment \cite{brockman2016openai} show that the proposed Hebbian AIF approach outperforms the use of Q-learning and compares favorably to the backprop-trained Deep AIF system of \cite{ozan}, while \textit{not requiring} any replay buffer, as in typical reinforcement learning systems \cite{sutton}. Our derivations and experiments add to a growing number of work addressing the study of Hebbian Active Inference \cite{faitc1}, \cite{faitc2}.

This paper is organized as follows. Background theory about Hebbian learning networks is provided in Section \ref{baselinenet}. Our Hebbian AIF methods are covered in Section \ref{methods}. Experimental results are shown in Section \ref{exps}. Conclusions are provided in Section \ref{conclusion}.  

\section{Background Theory on Hebbian Learning Networks}
\label{baselinenet}
Inspired by previous works that model the neural activity of biological agents through Sparse Coding \cite{fristonpredic}, \cite{9892362} (such as in the \textit{mushroom body} of an insect's brain \cite{liang2021can}), we model each individual Hebbian Ensemble layer of our networks as an identically-distributed Gaussian likelihood model with a Laplacian prior on the neural activity $c$:
\begin{equation}
    p(c|o,\Phi) \sim \exp{(-|| \Phi c - o ||_2^2)} \exp{(-\lambda ||c||_1)}
    \label{sparsemodel}
\end{equation}
where $o$ is the input of dimension $N$, $c$ is the output of dimension $M$, $\Phi$ is the $N \times M$ weight matrix of the layer (also called \textit{dictionary}), and $\lambda$ is a hyper-parameter setting the scale of the Laplacian prior. Choosing a Laplacian prior is motivated by the fact that it promotes sparsity in the output neural code in a way similar to how sparsity is induced in networks of Spiking Leaky Integrate-and-Fire neurons, modelling cortical neural activity \cite{9892362}. 

Under Sparse Coding (\ref{sparsemodel}), inference of $c$ and learning of $\Phi$ is carried via \cite{9892362}:
\begin{equation}
    C, \Phi = \arg \min_{C, \Phi} \sum_l  || \Phi c_l - o_l ||_2^2 + \lambda ||c_l||_1 \textbf{ with } C = \{c_l, \forall l\}
    \label{dlbp}
\end{equation}
which can be solved via Proximal Stochastic Gradient Descent \cite{NEURIPS2019d073bb8d}, by \textit{alternating} between: \textit{a)} the inference of $c_l$, given the current input $o_l$ and the weight $\Phi$ and \textit{b)} the learning of $\Phi$, given the current $c_l$ and $o_l$. 

Hence, we instantiate Hebbian layers as the dynamical system given in (\ref{dynamics}), where $T$ denotes the transpose, $\eta_c$ is the coding rate, $\eta_d$ is the learning rate and $\textbf{Prox}_{\lambda ||.||_1}$ is the proximal operator to the $l_1$ norm (non-linearity) \cite{lin2018sparse}. For each input $o_i$, the neural and weight dynamics of the Hebbian network follows the update rules in (\ref{dynamics}) for an arbitrary number of iterations (set to $100$ in this work as a good balance between speed of convergence and convergence quality), in order to infer the corresponding $c_l$ and learn $\Phi$ \cite{9892362}. 
\begin{equation}
     \begin{cases}
     c_l \leftarrow \textbf{Prox}_{\lambda ||.||_1} \{ c_l - \eta_c\Phi^T (\Phi c_l - o_l)  \} \\ 
     \Phi \leftarrow \Phi - \eta_d (\Phi c_l - o_l) c_l^T
    \end{cases}   
    \label{dynamics}
\end{equation}
with $\textbf{Prox}_{\lambda ||.||_1}$ acting as the neural non-linearity:
\begin{equation}
\textbf{Prox}_{\lambda ||.||_1}(c_i) = \text{sign}(c_i) \max(0, |c_i| - \eta_c \lambda), \forall i 
    \label{proxop}
\end{equation}

\begin{figure}
\centering
\includegraphics[scale = 0.56]{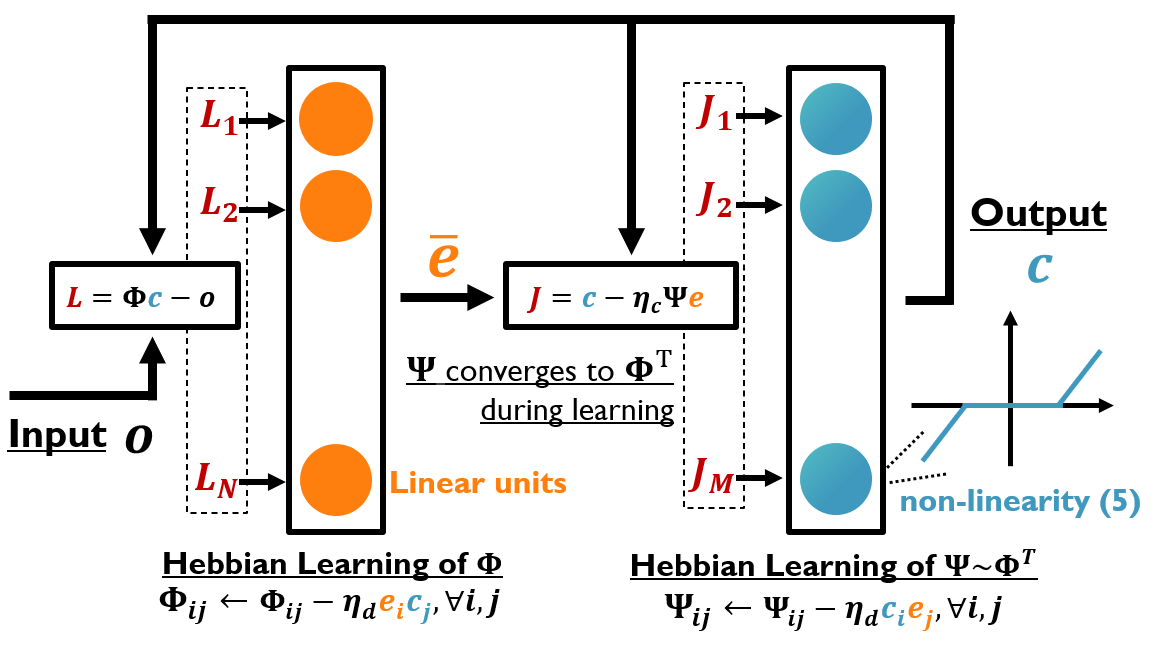}
\caption{\textit{\textbf{Baseline Hebbian network architecture used in this work.} The dynamics of the network follow (\ref{dynamics}) and minimize (\ref{dlbp}), given subsequent input vectors $o$. Each layer possesses its own weight matrix $\Phi, \Psi$ which evolve through Hebbian plasticity ($\Psi \sim \Phi^T$ in (\ref{dynamics}), as an \textit{independent}, local set of weights).}} 
\label{basenetarch}
\end{figure}

From a neural point of view, the dynamical system of (\ref{dynamics}) can be implemented as the network architecture in Fig. \ref{basenetarch}, where all weight updates follow the standard Hebbian rule (\ref{hebbianrule}) \cite{9892362}.


\section{Active Inference in Hebbian Learning Networks}
\label{methods}
In this Section, we show how the Hebbian network described above in Section \ref{baselinenet} is utilized in order to build an AIF system. First, we describe how Variational Free Energy minimization can be performed by a cascade of two Hebbian networks: a \textit{state-transition} network predicting the next latent states given the previous ones, and a \textit{posterior} network providing latent states given input observations. Crucially, it is shown that Free Energy minimization necessitates top-down Hebbian learning connections from the \textit{state-transition} network towards the \textit{posterior} network, steering the posterior output activity towards the state-transition output during learning. Then, we show how the Expected Free Energy is computed by generating state transition roll-outs. 
\subsection{Minimizing the Variational Free Energy}

The Variational Free Energy can be decomposed as \cite{ozan} \cite{fristnature} (where $\mathbb{E}$ denotes the expected value):
\begin{equation}
    \mathcal{F} = D_{KL}[q_{\Phi_P}(s_l|o_{l-1},a_{l-1})||p_{\Phi_S}(s_l|s_{l-1}, a_{l-1})] - \mathbb{E}_{q}[\log(p_{\Phi_L}(o_l|s_l))]
    \label{freeenergydec1}
\end{equation}
with the parametrized densities $q_{\Phi_P}, p_{\Phi_S}, p_{\Phi_L}$ described in Section \ref{intro}. Since the Hebbian network architecture used in this work intrinsically provides a mean to reconstruct its input $x_l$ in (\ref{dlbp}) (i.e., \textit{likelihood} modelling) from its produced latent code $c_l$ in (\ref{dlbp}) (i.e., \textit{posterior} modelling), using the \textit{same} \textit{dictionary} parameter matrix $\Phi$ in (\ref{dlbp}) that was used to generate $c_l$ via (\ref{dynamics}), we have $\Phi_L = \Phi_P$ in (\ref{freeenergydec1}) and we will solely use $\Phi_P$ below to denote the \textit{posterior} weight matrix.

Under the assumption of Gaussian likelihood with identity covariance in (\ref{sparsemodel}), the KL divergence $D_{KL}$ in $\mathcal{F}$ can be simplified to \cite{kldiv}:

\begin{equation}
    \mathcal{F} \sim || \Phi_S c_{S,l} - \{s_l(\Phi_P), a_l\} ||_2^2 + || \Phi_P s_l - \{o_l,a_l\} ||_2^2
    \label{freeenergydec1step2}
\end{equation}
where $s_l(\Phi_P)$ explicits the dependency of $s_l$ on $\Phi_P$ ($s_l$ being the posterior network output activity) and $c_S$ denotes the output activity of the state-transition network given $s_l$. The Free Energy in (\ref{freeenergydec1step2}) must be minimized with regard to the state transition weights $\Phi_S$ and the posterior weights $\Phi_P$ during learning:
\begin{equation}
    \Phi_S, \Phi_P = \arg  \min_{\Phi_S, \Phi_P} || \Phi_S c_{S,l} - \{s_l(\Phi_P), a_l\} ||_2^2 + || \Phi_P s_l - \{o_l,a_l\} ||_2^2,  \forall l
    \label{freeenergymin1}
\end{equation}
This indicates that it is not only the state transition model that must be steered towards the posterior model, but also, the posterior model must be steered towards the output of the state-transition network. This effect can be achieved by re-formulating the optimization in (\ref{freeenergymin1}) as:
\begin{equation}
\begin{cases}
    \Phi_S = \arg  \min_{\Phi_S} || \Phi_S c_{S,l} - \{s_l, a_l\}  ||_2^2, \forall l & \text{(a)} \\
    \Phi_P = \arg  \min_{\Phi_P} || \Phi_P s_l - \{o_l,a_l\} ||_2^2 + || \Phi_P (\Phi_S c_{S,l}) - \{o_l,a_l\} ||_2^2 & \text{(b)} 
\end{cases}
    \label{freeenergymin}
\end{equation}

Intuitively, the right-hand term in (\ref{freeenergymin} b) steers the \textit{posterior} model towards the \textit{state-transition} model by first re-projecting the output activity of the state-transition network $c_{S}$ into the latent space as $\Phi_S c_{S}$ (considering $\Phi_S$ fixed). Then, minimizing $|| \Phi_P (\Phi_S c_{S,l}) - \{o_l,a_l\} ||_2^2$ modifies $\Phi_P$ in order to steer its posterior output $s_l$ towards the re-projected state-transition activity $\Phi_S c_{S}$ (considering $\{o_l,a_l\}$ fixed).

\subsubsection{State-Transition Model}
\begin{figure}
\centering
\includegraphics[scale = 0.7]{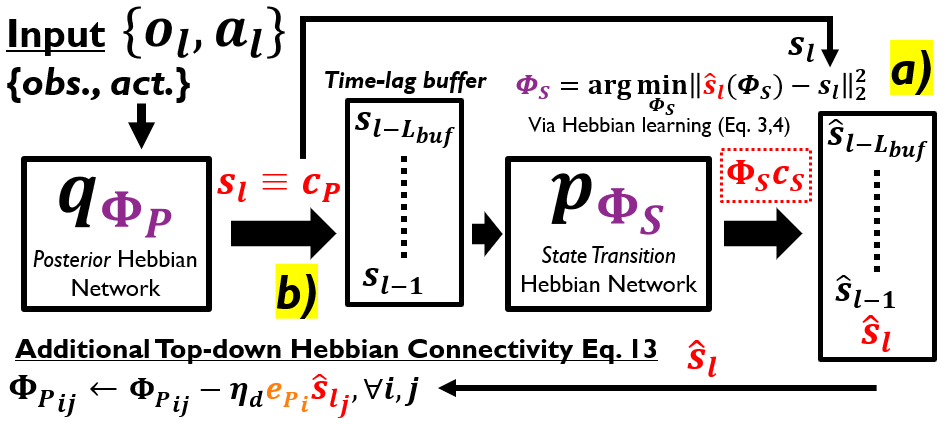}
\caption{\textit{\textbf{Hebbian Active Inference Architecture.} a) The \textit{state-transition} network takes as input the $L_{buf}$ previous latent states produced by the \textit{posterior} network, and projects them onto its internal representation $c_{S}$ via the learned $\Phi_S$. When producing roll-outs, the \textit{state-transition} network estimates the next state $\hat{s}_l$ by re-projecting the output activity $c_{S}$ due to $[s_{l-L_{buf}}, \hdots, s_{l-1}]$  back onto the input space via (\ref{statetrans}). b) The \textit{posterior} network takes observation-action pairs as input and produces latent states $s$ (corresponding to the output in Fig. \ref{basenetarch}). In addition to the Hebbian mechanisms depicted in Fig. \ref{basenetarch}, the weights $\Phi_P$ of the \textit{posterior} network are also subject to a \textit{top-down} Hebbian learning mechanism for minimizing the second term in (\ref{freeenergymin} b).}} 
\label{sysarch}
\end{figure}

Inspired by prior work on dictionary-based sequence modeling \cite{inspiration}, we implement the transition model $p_{\Phi_S}(s_l|\Tilde{s}_{l-1}, \Tilde{a}_{l-1})$ as an \textit{auto-regressive} Hebbian network (see Fig. \ref{sysarch} a), taking as input a sequence of state-and-action \textit{history} $\Tilde{s}_{l-1} = [s_{l-1}, \hdots, s_{l-L_{buf}}]$, $\Tilde{a}_{l-1} = [a_{l-1}, \hdots, a_{l-L_{buf}}]$ and inferring the next state  $\Tilde{s}_{l} = [s_{l}, \hdots, s_{l-L_{buf}}]$ as the re-projection of its internal sparse code $c_{S}$ in the input space through the network weights $\Phi_S$:
\begin{equation}
     \Tilde{s}_{l} = \Phi_S c_{S,l} 
    \label{statetrans}
\end{equation}
where $\Phi_{S,j}$ and $c_{S, j}$ respectively denote the weight vector and the sparse code of each layer $j$ in the state transition network. Therefore, the state-transition network effectively projects the $L_{buf}$ previous states (noted $\Tilde{s}_{l-1}$) into a common internal sparse code $c_{S}$ and reconstructs the next states $\Tilde{s}_{l}$ by re-projection of $c_{S}$ into the input space.

The state-transition network learns its weights $\Phi_S$ following (\ref{freeenergymin} a) and infers its output activity $c_{S}$ via sparse coding (see Section \ref{baselinenet}):
\begin{equation}
    \Phi_S, c_{S,l}  = \arg  \min_{\Phi_S, c_{S,l} } || \Phi_S c_{S,l} - \Tilde{s}_{l} ||_2^2 + \lambda_P || c_{S,l}||_1, \forall l 
    \label{statetranlearn}
\end{equation}
where $\lambda_P$ is a parameter that sets the strength of the \textit{sparsity} of the \textit{state-transition} output activity. (\ref{statetranlearn}) can therefore be implemented via the Sparse Coding-based Hebbian learning ensemble described in Section \ref{baselinenet}. This auto-regressive strategy enables the network to learn state predictions using Hebbian learning, \textit{without} the need for non-bio-plausible back-propagation through time (BPTT) \cite{inspiration}, \cite{bptt}.

In order to prevent the vanishing or exploding of the state transition model when producing roll-outs further in time, we regularize the norm of the reconstructed states $s_{l}$ to an arbitrary magnitude $\alpha$ using (\ref{homeostasis}). We keep $\alpha = 5$ in our experiments in Section \ref{exps}, giving a good balance for the dynamic range of the network output activity (adjusted empirically).
\begin{equation}
    s_{l} = \alpha \frac{s_{l}}{||s_{l}||_2}
    \label{homeostasis}
\end{equation}
\subsubsection{Posterior Model}
 Similar to the state-transition model, we use a Hebbian ensemble as posterior model, where the internal sparse code $c_{P}$ (see Fig. \ref{sysarch} b) is identified as the hidden state $s_{l} \equiv c_{P,l}$ inferred by the posterior network $q_{\nu}(s_l|o_{l-1},a_{l-1})$, given the observation and action pair $\{o_{l-1}, a_{l-1}\}$ in (\ref{sparsemodel}). 

 Therefore, the posterior network learns its weights $\Phi_P$ following (\ref{freeenergymin} b) and infers its output activity $s_{l} = c_{P,l}$ via sparse coding (see Section \ref{baselinenet}):
\begin{equation}
    \Phi_P, c_{P,l}  = \arg  \min_{\Phi_P, c_{P,l} } \underbrace{|| \Phi_P c_{P,l} - \{o_l,a_l\} ||_2^2 + \lambda_Q ||c_{P,l}||_1}_\text{Standard Sparse Coding} + \overbrace{|| \Phi_P (\Phi_S c_{S,l}) - \{o_l,a_l\} ||_2^2}^\text{\textbf{Top-down Connection}} 
    \label{postlearn}
\end{equation}
for all $l$, where $\lambda_Q$ sets the strength of the \textit{sparsity} of the \textit{posterior} output activity. The left-hand \textit{standard sparse coding} term in (\ref{postlearn}) can be implemented via the Hebbian learning ensemble described in Section \ref{baselinenet}, while the right-hand term in (\ref{postlearn}) can be implemented using \textit{top-down} connections from the state-transition output activity $c_{S}$ towards the posterior network, via the state-transition weight matrix $\Phi_S$ (see Fig. \ref{sysarch} b).

Finally, here again, we apply the homeostasis rule (\ref{homeostasis}) to the inferred posterior state, effectively constraining $s_{l}$ to lie on the $\alpha$-sphere manifold.



\subsection{Minimizing the Expected Free Energy}

Given a policy $\pi$, the Expected Free Energy $G(\pi)$ (EFE) can be written as \cite{10.3389/fpsyg.2013.00710}:
\begin{multline}
    G(\pi) = \sum_{l} \mathbb{E}_{q(o_l,s_l|\pi)}[\log q(s_l|\pi) - \log p(s_l,o_l|\pi)] \\ = \sum_{l} -H\{q(s_l|\pi)\} - \mathbb{E}_{q(o_l,s_l|\pi)}[\log p(s_l,o_l|\pi)] 
    \label{EFE}
\end{multline}
 where $H$ denotes the Shannon entropy. It can be seen in (\ref{EFE}) that selecting a policy that minimizes the EFE entails the maximization of the posterior entropy  (promoting exploration) and the joint posterior over the states and observations  (reaching the desired goal) \cite{10.3389/fpsyg.2013.00710}. 

 In order to reach the desired goal, we produce roll-outs of states $s_l$ given a certain policy and approximate the term $- \mathbb{E}_{q}[\log p(s_l,o_l|\pi)]$ to be minimized as:
\begin{equation}
    - \mathbb{E}_{q(o_l,s_l|\pi)}[\log p(s_l,o_l|\pi)] \sim ||s_l - s^*||_2^2
    \label{goalterm}
\end{equation}
where $s^*$is the desired state that the agent must reach, corresponding to a desired observation (e.g., the agent's position). Since the observation $o_l$ can encompass more than just the goal to be reached (i.e., the observation $o_l$ could be both the position and the velocity of an agent, even though the desired goal is to reach a specific position regardless of the velocity), we compute the desired \textit{goal} state $s^*$ as:
\begin{equation}
    s^* = \arg \max_{s} \int_{\omega \in D_\omega} q(s|\Omega^*,\omega) d\omega
    \label{desiredstate}
\end{equation}
where $\Omega^*$ contains all observations that must be reached in order to attain the desired goal and $\omega$ designates all observation modalities that are \textit{not} taking part in defining the goal that must be reached, with $D_\omega$ their domain of definition. In practice, (\ref{desiredstate}) is estimated by averaging the output of the posterior network, while sweeping $\omega$ for a grid of possible values and keeping $\Omega^*$ fixed. 

Regarding the exploration term in (\ref{EFE}), our Hebbian network does not directly allow the estimation of the entropy $H\{q(o_l,s_l|\pi)\}$, since the network does not infer standard deviations as in a variational auto-encoder (VAE) \cite{ozan}. We propose to replace the maximization of the entropy $H\{q(o_l,s_l|\pi)\}$ with a surrogate term, crafted to promote exploration as well. As a surrogate for $H\{q(o_l,s_l|\pi)\}$, we choose to maximize the variance (noted $\text{Var}$) of the state trajectory $s_l, \forall l = 1,...,L$ along time during the roll-outs. Intuitively, a state trajectory that presents lots of variation in time will promote the exploration of new states, providing a similar qualitative effect as maximizing $H\{q(o_l,s_l|\pi)\}$. Therefore, we select the policy $\pi$ such that the distance to the desired state is minimized, while achieving a state trajectory variance larger than a certain threshold $t_v$.
\begin{equation}
    \pi^* = \arg \min_{\pi} \mathcal{G}(\pi) = \sum_{l=1}^{L} ||s_l - s^*||_2^2 \hspace{10pt} \text{ s.t. } \hspace{10pt} \text{Var}(||s_l - s^*||_2^2, l=1,...,L) \geq t_v
    \label{EFEest}
\end{equation}

Given a set of $N_p$ policies to try, $t_v$ can be determined in an adaptive way as:
\begin{equation}
    t_v = \beta \times \frac{1}{2} [\boldsymbol{\max_{\pi}}(\text{Var}(||s_l(\pi) - s^*||_2^2, \forall l)) + \boldsymbol{\min_{\pi}}(\text{Var}(||s_l(\pi) - s^*||_2^2, \forall l))] 
    \label{tvestadaptive}
\end{equation}
where $\beta$ is the strength hyper-parameter (empirically set to $0.5$ in our experiments reported below).

\section{Experimental Results}
\label{exps}
The aim of our experimental studies is to determine \textit{i)} how the main network hyper-parameters (number of neurons, sparsity in output activity,...) impact the success rate of the proposed Hebbian AIF system; \textit{ii)} to what extent Hebbian AIF is robust when learning without using a replay buffer and \textit{iii)} how Hebbian AIF compares to Q-learning (which uses \textit{dense rewards} versus \textit{unsupervised learning} in Hebbian AIF).

\subsection{Mountain Car Environment}
We perform experiments in the \textit{Mountain Car} environment from the OpenAI gym suite \cite{brockman2016openai}. In this task, a car starts at a \textit{random} position at the bottom of a hill and is expected to reach the top of a mountain within $200$ time steps. The agent is subject to gravity and cannot reach the goal trivially, just by accelerating towards it. Rather, the agent must learn to gain momentum before accelerating towards the goal.

In this environment, the x-axis position $x$ and the velocity $v_x$ of the car constitute the input observations to the Hebbian AIF network. Before feeding the observation tuple $(x, v_x)$ to our Hebbian network, we normalize $(x, v_x)$ using (\ref{equalizi}) in order to equalize the dynamic range of the position and velocity signals:
\begin{equation}
    \begin{cases}
    x \xleftarrow{} \frac{x - \mu_x}{\sigma_x} \\
    v_x \xleftarrow{} \frac{x - \mu_{v_x}}{\sigma_{v_x}}
    \end{cases}
    \label{equalizi}
\end{equation}
where $(\mu_x, \sigma_x)$ and $(\mu_{v_x}, \sigma_{v_x})$ denote the mean and standard deviation of the position and velocity signals respectively (estimated during random environment runs).

We use an action space constituted by two discrete actions: \textit{accelerate to the left} and \textit{accelerate to the right}. In addition, each action is repeated for $10$ consecutive time steps once selected during the Expected Free Energy minimization in (\ref{EFEest}).

In order to compute the Expected Free Energy, we generate roll-outs of $L = 200$ time step predictions for $100$ different random policies $\pi^j, j=1,...100$ with equal probability of selecting the \textit{accelerate to the left} or the \textit{accelerate to the right} actions.

As learning rate for the Hebbian learning mechanism (\ref{dynamics}), we use $\eta_d=10^{-4}$ with a \textit{decay rate} of $0.8$ applied at the end of each \textit{successful} episode, i.e. if the episode terminates successfully, $\eta_d \xleftarrow{} \eta_d \times 0.8$ (else no decay is applied on $\eta_d$). All weights are initialized randomly from a normal distribution with standard deviation $0.01$.

In the remainder of this Sections, we perform all our experiments using a 10-fold validation approach, by reporting the success rate curves as averages over 10 different runs (with $35$ episodes per runs), with different random network initializations. For each run, we compute the success rate curve using a moving average window of size $5$, and report the mean success rate curve by averaging over the $10$ runs, alongside with its standard deviation (see e.g. Fig. \ref{postneursweep}). We will now study the impact of the various network hyper-parameters on the achieved success rates.

\subsection{Impact of the Number of Neurons in the Posterior and State Transition Networks}

\begin{figure}
\centering
\includegraphics[scale = 0.63]{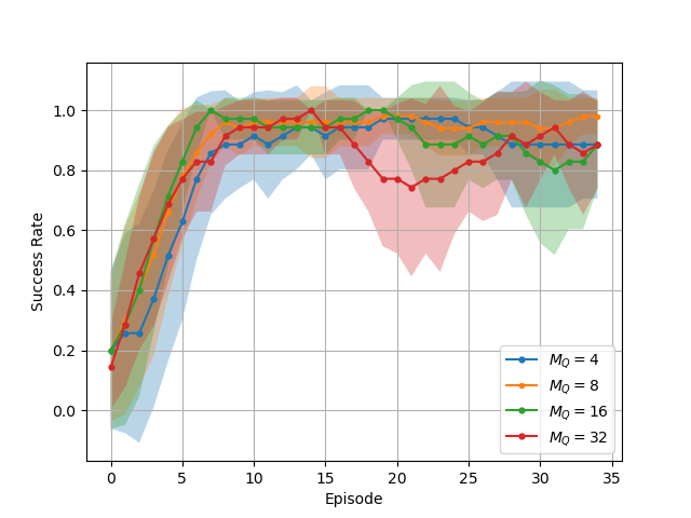}
\caption{\textbf{Impact on the success rate when changing the \textit{number of neurons $M_Q$} in the \textit{posterior network}.}} 
\label{postneursweep}
\end{figure}

\begin{figure}
\centering
\includegraphics[scale = 0.6]{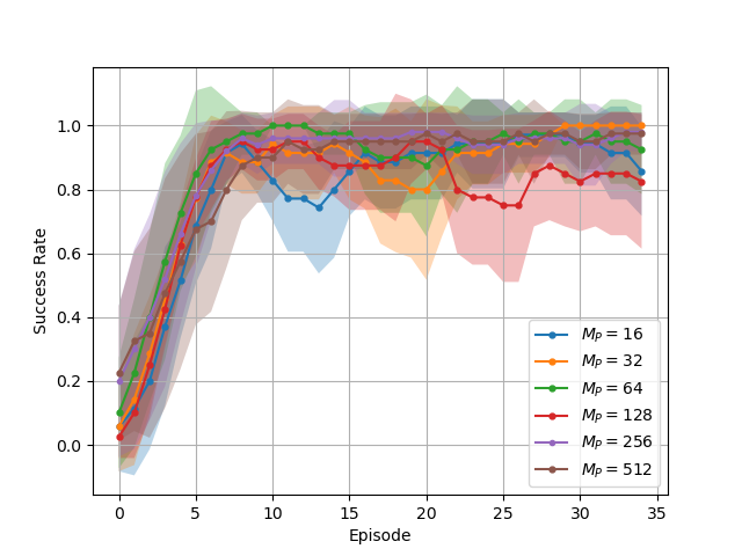}
\caption{\textbf{Impact on the success rate when changing the \textit{number of neurons $M_P$} in the \textit{state transition network}.}} 
\label{statetransneursweep}
\end{figure}

Fig. \ref{postneursweep} and \ref{statetransneursweep} show the effect of sweeping the number of coding neurons $M_Q$ and $M_P$ in both the \textit{posterior} and \textit{state-transition} networks. Fig. \ref{postneursweep} shows that for $M_Q < 8$, the success rate is sub-optimal, but reaches a steady plateau around  $M_Q = 8$ (orange curve in Fig. \ref{postneursweep}). Then, as $M_Q$ is increased for $M_Q > 8$, the success rate becomes sub-optimal again, with dips in the performance along the episodes (e.g., red curve in Fig. \ref{postneursweep}). This phenomenon can be explained as follows: for $M_Q < 8$, the posterior network does not have enough parameters to capture the input dynamics into its latent space and \textit{under-fits}, while for $M_Q > 8$, the posterior network starts over-fitting, reducing the success rate again.

Regarding the \textit{state-transition} network, Fig. \ref{statetransneursweep} shows that the higher the number of neurons $M_P$, the flatter the success rate curves become, leading to higher performance. The state transition network does not seem to over-fit as $M_P$ is increased (for $\lambda_P=10^{-4}$ kept fixed). Rather, Fig. \ref{statetransneursweep} indicates that a higher state-transition network capacity is beneficial for capturing important dynamics in the latent space, at the output of the posterior network.

\subsection{Impact of the Sparsity of the Output Activity in the Posterior and State Transition Networks}

\begin{figure}[htbp]
\centering
\includegraphics[scale = 0.545]{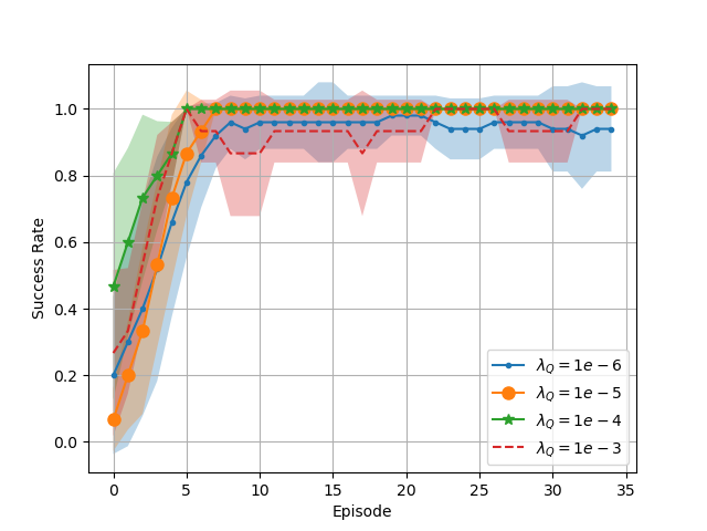}
\caption{\textbf{Impact on the success rate when changing the \textit{sparsity hyper-parameter $\lambda_Q$} in the \textit{posterior network}.}} 
\label{postspars}
\end{figure}

\begin{figure}[htbp]
\centering
\includegraphics[scale = 0.68]{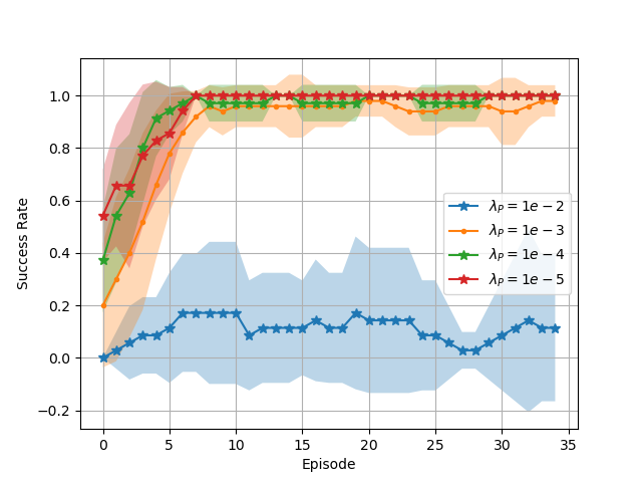}
\caption{\textbf{Impact on the success rate when changing the \textit{sparsity $\lambda_P$} in the \textit{state transition network}.}} 
\label{statetranssparsity}
\end{figure}

Fig. \ref{postspars} and \ref{statetranssparsity} show the effect of sweeping the \textit{sparsity-defining} hyper-parameters $\lambda_Q$ and $\lambda_P$ in both the \textit{posterior} and \textit{state-transition} networks. For the posterior network, Fig. \ref{postspars} shows that the success rate performance initially grows as $\lambda_Q$ is increased from $\lambda_Q=10^{-6}$ to $\lambda_Q=10^{-5}$. Doing so, the non-linearity of the posterior network is increased, better capturing observation features into its latent space. Then, as $\lambda_Q$ grows past $\lambda_Q=10^{-4}$, the success rate degrades again, indicating a too strong posterior network non-linearity. 

Regarding the \textit{state-transition} network, Fig. \ref{statetranssparsity} shows that the lower $\lambda_P$, the higher the success rate becomes. This suggests that making the state-transition network \textit{more linear} (i.e., lower $\lambda_P$) better captures the dynamics of the latent space produced by the posterior network (other parameters kept fixed).

\subsection{Impact of the Time-Lag Buffer Length on Task Performance}

\begin{figure}
\centering
\includegraphics[scale = 0.6]{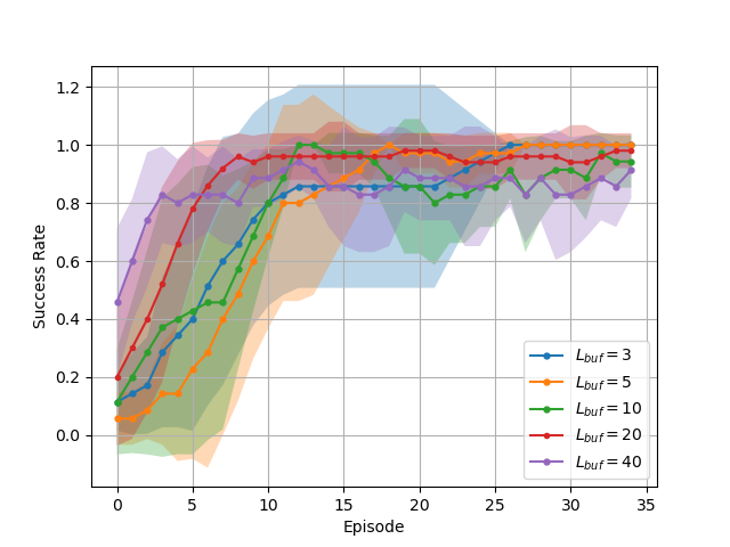}
\caption{\textbf{Success rate when changing the \textit{time-lag buffer length $L_{buf}$}.}} 
\label{timelag}
\end{figure}

Fig. \ref{timelag} shows how the length $L_{buf}$ of the time-lag buffer impacts the achieved success rate. Initially, as $L_{buf}$ increases, the success rate increases as well, due to an increased availability of past latent states used by the state-transition network to estimate the next expected state. Then, as $L_{buf}$ is further increased for $L_{buf} > 20$, the success rate drops again due to the addition of latent states \textit{from deep in the past} that are less useful for estimating the present dynamics.

\subsection{Comparing Hebbian AIF against the Use of a Replay Buffer and against Q-learning}
\begin{figure}
\centering
\includegraphics[scale = 0.58]{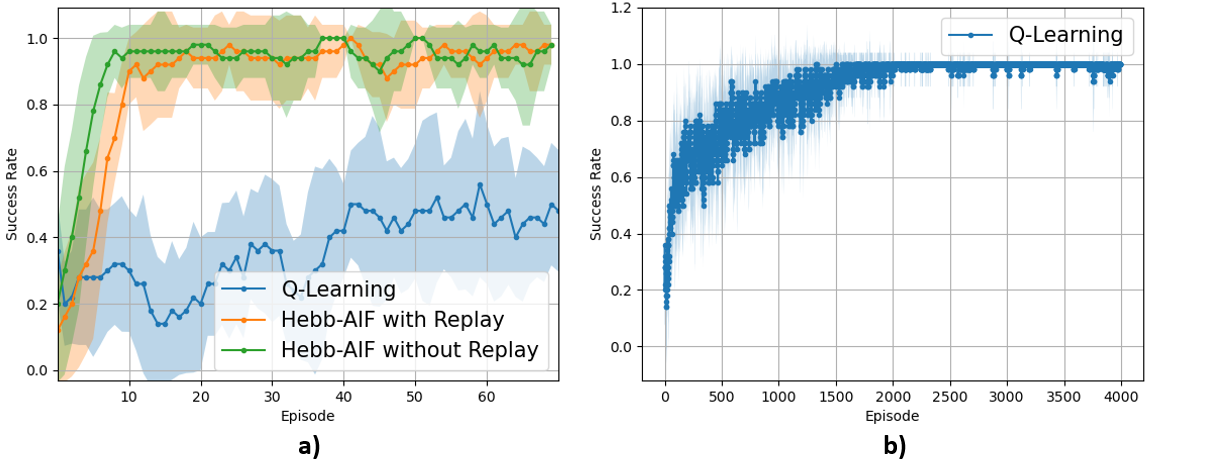}
\caption{\textbf{Hebbian AIF versus the use of a \textit{replay buffer} and \textit{Q-learning} (a).} Q-learning needs two orders of magnitude more episodes in order to converge (b).} 
\label{comptoqlearn}
\end{figure}

Fig. \ref{comptoqlearn} compares the success rate obtained using our proposed Hebbian AIF system against \textit{a)} the use of a replay buffer during learning and \textit{b)} the use of a Q-learning agent. Experience replay is done by saving the history of observation-action pairs in a buffer after each episode. After the end of the episode, a past experience is randomly selected and used to train the Hebbian AIF system for one episode.

Regarding the Q-learning setup, we use a standard Q-table learning approach \cite{sutton}, with the python implementation proposed in \cite{urlcode}. 

Fig. \ref{comptoqlearn} shows that our Hebbian AIF system converges much faster than the Q-learning system and behaves in a comparable manner to the Hebbian AIF setup with a \textit{replay buffer}. Indeed, the Q-learning agent needs much more episodes in order to converge, despite the fact that it utilizes the \textit{dense rewards} provided by the Mountain Car environment \cite{brockman2016openai} (versus \textit{unsupervised} learning in the case of Hebbian AIF). This confirms prior observations about the efficient convergence of AIF systems, due to their ability to learn a generative model of environment dynamics (versus supervised learning of a Q-table) \cite{ozan}. 

Finally, it is interesting to note that, compared to the Deep AIF results \textit{reported in} \cite{ozan} (using a fully-connected 2-hidden-layer network trained through backprop), the Hebbian AIF system proposed in this work 
eventually reaches $\sim 100\%$ success rate (see red curve in Fig. \ref{statetranssparsity}) while the system in \cite{ozan} reaches $\sim 95\%$, motivating further investigations of Hebbian learning for AIF systems.
\section{Conclusion}
\label{conclusion}
This paper has investigated how neural ensembles equipped with local Hebbian plasticity can perform active inference for the control of dynamical agents. First, a Hebbian network architecture performing joint dictionary learning and sparse coding has been introduced for implementing both the posterior and the state-transition models forming our generative Active Inference system. Then, it has been shown how Free Energy minimization can be performed by the proposed Hebbian AIF system. 
Finally, extensive experiments for parameter exploration and benchmarking have been performed to study the impact of the network parameters on the task performance. Experimental results on the Mountain Car environment show that the proposed system outperforms the use of Q-learning, while \textit{not requiring} the use of a replay buffer during learning, motivating \textit{future investigations} of using Hebbian learning for designing active inference systems. 

\section*{Acknowledgement}
This research was partially funded by a Long Stay Abroad grant from the Flemish Fund of Research - Fonds Wetenschappelijk Onderzoek (FWO) - grant V413023N. This research received funding from the Flemish Government under the “Onderzoeksprogramma Artificiële Intelligentie (AI) Vlaanderen” programme.

\end{document}